\documentclass{article}

\usepackage[main, final]{neurips_2026}

\usepackage[utf8]{inputenc} 
\usepackage[T1]{fontenc}    
\usepackage{hyperref}       
\usepackage{url}            
\usepackage{booktabs}       
\usepackage{amsfonts}       
\usepackage{nicefrac}       
\usepackage{microtype}      
\usepackage{xcolor}         
\usepackage[pdftex]{graphicx}
\usepackage{amsmath}
\usepackage{bm}
\usepackage{wrapfig}
\usepackage{subcaption}

\makeatletter
\renewcommand{\paragraph}{%
  \vspace{-1mm}
  \@startsection{paragraph}{4}{\z@}%
  {0.2ex}
  {-1em}
  {\normalfont\normalsize\bfseries}%
}
\makeatother

\title{What-Where Transformer: A Slot-Centric Visual Backbone for Concurrent Representation and Localization}

\author{%
  Ryota Yoshihashi\textsuperscript{1} \\
  \And
  \hspace{-2mm}Masahiro Kada\textsuperscript{1}\\
  \And
  \hspace{-2mm}Satoshi Ikehata\textsuperscript{2,}\textsuperscript{3}\\
  \And
  \hspace{-2mm}Rei Kawakami\textsuperscript{1}\\
  \And
  \hspace{-2mm}Ikuro Sato\textsuperscript{1,}\textsuperscript{2}\\
  \And \vspace{-10mm}
  \\ 
  \textsuperscript{1}Institute of Science Tokyo \\
  \textsuperscript{2}DENSO IT Laboratory \\
  \textsuperscript{3}National Institute of Informatics \\
}

\begin{document}

\maketitle

\vspace{-6mm}
\begin{abstract}\vspace{-4mm}
  Many image understanding tasks involve identifying {\it what} is present and {\it where} it appears. 
  However, localization tasks for {\it where}, such as object discovery, detection, and segmentation, are often more complex than classification for {\it what}.
 One possible reason is that classification-oriented backbones tend to emphasize semantic {\it what}, while implicitly entangling or suppressing {\it where}.
  We focus on what?where separation, an inductive bias that encourages models to decompose object representations and their location.
  To incorporate this bias throughout an attentive backbone in the style of ViT, we propose the What-Where Transformer (WWT). Our method introduces two key novel designs:
(1) it treats tokens as representations of {\it what} and attention maps as representations of {\it where}, and processes them in concurrent feed-forward modules via a multi-stream, slot-based architecture;
(2) it reuses both the final-layer tokens and attention maps for downstream tasks, and directly exposes them to gradients derived from task losses, facilitating explicit learning of localization.
We demonstrate that even under standard single-label classification-based supervision on ImageNet, WWT exhibits emergent multiple object discovery directly from raw attention maps, rather than via additional postprocessing such as token clustering. Furthermore, WWT achieves superior performance compared to ViT-based methods on zero-shot object discovery and weakly supervised semantic segmentation, and it is transferable to various localization setups with minimal modifications. Code will be published after acceptance.
\end{abstract}

\vspace{-9mm} \section{Introduction}\vspace{-4mm} 
Transformer equipped with attention mechanisms \cite{vaswani2017attention} has demonstrated strong performance in image recognition. 
The Vision Transformer (ViT) \cite{dosovitskiy2020image}, used for classification and feature extraction, and the Detection Transformer (DETR) \cite{carion2020end}, specialized for object detection, along with their variants, are well known. Nevertheless, tasks beyond classification still typically require task-specific decoder architectures, such as those for detection or segmentation. This requirement increases the complexity of building unified visual systems that span multiple tasks.
In contrast, embedding localization ability directly into the visual encoder could provide more flexible and transferable localization for downstream spatial reasoning.

Conceptually, the attention maps in a ViT encoder should capture object regions, and some studies have partially observed such behavior \cite{caron2021emerging,rawlekar2026finding}.
However, in practice, encoder attention often suffers from noisy activations, vague boundaries, and limited semantic interpretability \cite{wang2022self,psomas2023keep,darcet2024vision}. 
As a result, these attention maps are not easily transferable to downstream tasks.
Consequently, many approaches decouple task decoders from encoder attention; the use of ViT encoder attention in downstream tasks is mostly limited to specific settings such as weakly supervised learning \cite{xu2022multi,gupta2022vitol}.

\begin{wrapfigure}{r}{0.55\textwidth}
    \centering
    \vspace{-0mm}\includegraphics[width=1.0\linewidth]{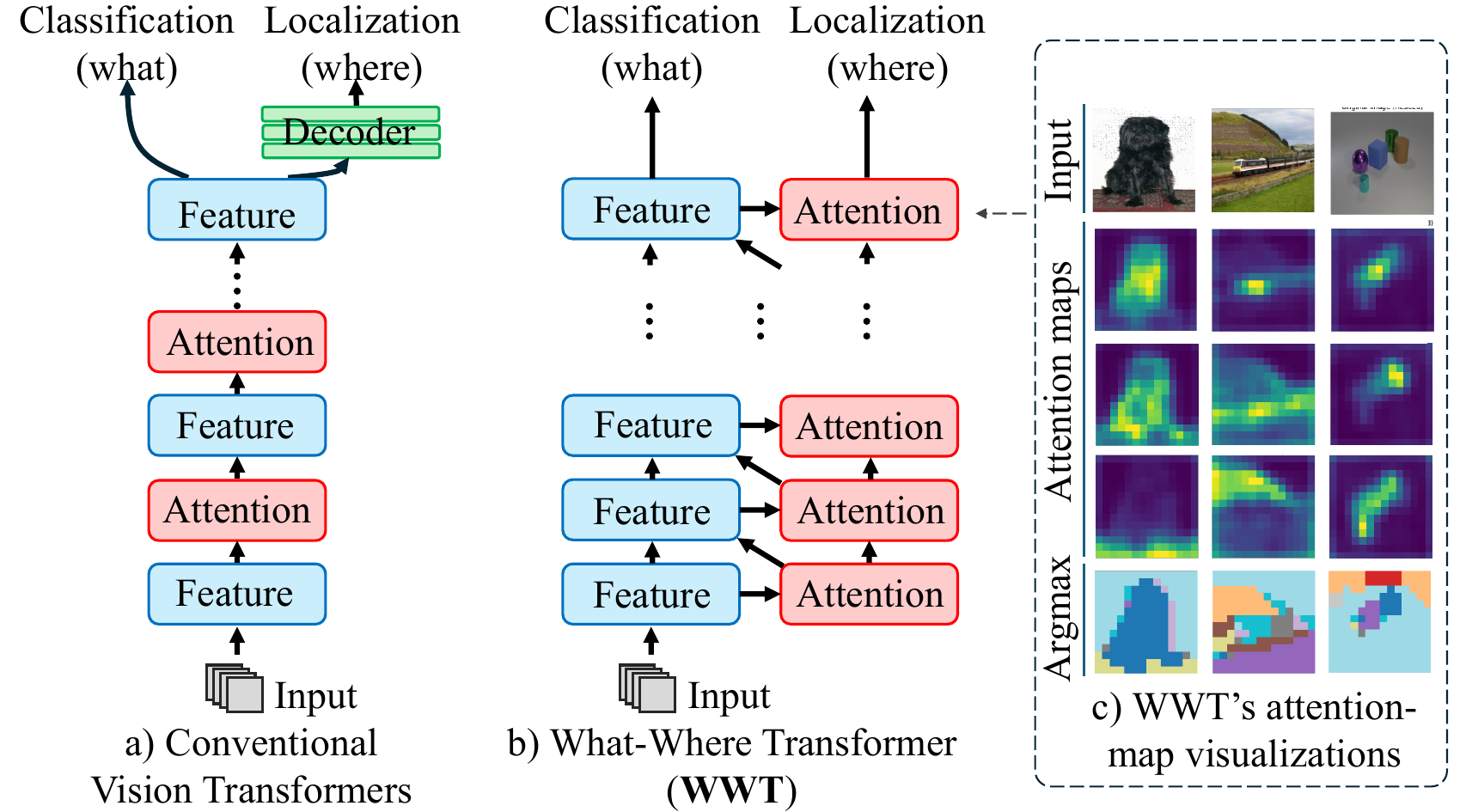}
    \vspace{-6mm}   
    \caption{Concept-level illustrations of a) conventional Vision Transformers and b) proposed What-Where Transformer (WWT). WWT features multi-stream design separating what pathway and where pathway. This design facilitates emerging separation of representations and their locations; as a result, c) it learns multiple-object discovery in the attention maps even in single-label classification tasks.}
    \vspace{-3mm}
    \label{fig:concept}
\end{wrapfigure}

We assume that the token-centric design is one of the factors contributing to the non-object-centric attentions in ViTs. 
ViT tokens, mostly derived from image-patch embeddings, serve as inputs, outputs, and internal states. Although attention maps temporarily capture spatial relationships between tokens, their role is limited to assisting token updates with spatial integration, and they are immediately absorbed back into the tokens within each block (Fig. \ref{fig:concept}a).
In contrast, this work draws inspiration from the concept of what?where separation \cite{mishkin1983object, gregor2010emergence}, in which the semantic content and spatial location of each visual entity are represented separately. 
To revive this classical concept in modern ViT, we introduce a design that pairs conventional token-like distributed representations with additional spatial representations, encouraging the model to associate them with ``what'' and ``where,'' respectively. 
By incorporating what?where separation as an inductive bias \cite{goyal2022inductive} in the architecture and loss, the model can acquire explicit positional representations even under pretraining tasks that do not directly require location information. As a result, not only ``what'' but also ``where'' can be extracted from the output, which is expected to simplify task-specific decoders and improve the transferability of the learned models as backbones for downstream tasks.

As an instantiation of what-where separation in ViT-style attentive vision architectures, we propose novel What-Where Transformer (WWT), illustrated in Fig. \ref{fig:concept}b). 
The model maintains separate representations for object semantics (``what'') and their spatial information (``where''), while enabling their concurrent interaction.
Each object in an image is represented as a pair consisting of a feature vector and a soft mask. The network is constructed by repeatedly stacking multi-stream feedforward layers, which process these pairs individually, and inter-attention layers, which allow communication between streams.
The soft mask not only aggregates spatial features like attention maps, but can also be directly interpreted as a region mask in the output. With relatively simple post-processing, it can be converted into boxes or dense labels, making it broadly applicable to localization tasks.

In the experiments, WWT was pretrained in a supervised manner on ImageNet-1k \cite{deng2009imagenet}, and we confirmed the following:
1) In ImageNet classification, it achieves performance comparable to conventional ViT while producing output masks with higher interpretability than ViT attention maps;
2) The output masks obtained from classification training can be directly applied to segmentation without retraining or additional modules.
3) It can also be extended to single- and multi-object discovery and supervised object detection with minimalistic modifications.
Especially, multi-object discovery solely from attention remains challenging for existing ViTs, since class-token attention often fails to separate multiple object regions, even when the objects belong to different classes.

Our contributions are summarized as follows.
1) Design principle: We frame the long-standing idea of what-where separation and analyze its role as an inductive bias, and establish our core design, tokens-as-representations and attentions-as-localization.
2) Architecture: We develop WWT, a ViT-style attentive backbone that instantiates the principle. Through mutual attention, WWT replaces ViT's dense token-token self-attention with slot-mediated bipartite communication and explicitly propagates localization states across layers.
3) Emergent multi-object discovery: Under our architectural inductive bias, we observed that multi-object-decomposed attentions emerge from single-label image-level classification training, and evaluated zero-shot object discovery tasks. 
4) Downstream versatility: Thanks to the localization mechanism embedded in the backbone, WWT can be applied to various localization tasks with minimal additional modules. Instead of pursuing per-task optimization, WWT provides a useful foundation for a compact generalist model.

\vspace{-3mm} \section{Related work}\vspace{-3mm} 
\paragraph{Transformers in vision}
Attention-based image backbones are predominantly built on Transformer encoders \cite{vaswani2017attention}, with the Vision Transformer (ViT) \cite{dosovitskiy2020image} serving as a standard image classifier and representation extractor.
While a plain Transformer can work effectively when trained on large-scale datasets \cite{dehghani2023scaling}, various adaptations have been proposed for vision tasks, incorporating multi-scale representations \cite{wang2021pyramid}, local attention \cite{liu2021swin}, and convolutional operations \cite{wu2021cvt}.
Dual-ViT \cite{yao2023dual} introduces learnable queries and a two-stream architecture.
An important common design shared by the existing major ViT variants is interleaved spatial and channel mixers~\cite{liu2022convnet,yu2023metaformer}, which are self-attention as spatial mixers and multi-layer perceptrons as channel mixers in ViT. 
While different choices of mixer modules (e.g., pooling, spatial MLPs \cite{liu2021pay,touvron2022resmlp}, or clustering \cite{ma2023image,deng2023perceptual}) have been explored, most approaches remain token-centric and rely on a largely single-stream processing paradigm.
In contrast, the key idea of our method---explicit factorization into what and where and multi-stream processing of them---explores a different direction in existing generalized ViT-based approaches.

Transformer as a visual task decoder forms another research line.
DETR \cite{carion2020end} for detection using a Transformer decoder and MaskFormer \cite{cheng2021per} for segmentation are also well known, and many variants \cite{zhu2020deformable,zhang2022dino,cheng2022masked} have been proposed.
On the other hand, there is a line of work that tackles detection and segmentation without relying on a strong decoder, instead using a Transformer encoder with lightweight task heads \cite{ranftl2021vision,li2022exploring}; this approach is particularly favored when integrating with large-scale vision?language models \cite{minderer2022simple,kirillov2023segment,li2025token} due to resource constraints.
Considering both directions, WWT incorporates cross-attention-like mechanisms and learnable queries, traditionally used in decoders, to adopt a backbone-centric design to address localization tasks.

\paragraph{Slot Attention}
The slot attention (SA) model is a type of autoencoder that aims to acquire ``slots'' as units of visual entities, achieving object-centric representation learning. The SA models use attention mechanisms to distribute and separate information from feature maps into slots, and they are characterized by their permutation invariance over slots \cite{kimura2024permutation}. As an application, it enables unsupervised segmentation by reconstructing images from each slot on an object-wise basis.
The original SA \cite{locatello2020object} adopts a convolutional encoder?decoder and an Attentive GRU \cite{chung2014empirical}. Subsequent works have replaced the GRU with a Transformer \cite{wu2023inverted}, enhanced the decoder with a Transformer \cite{singh2021illiterate} or a diffusion model \cite{jiang2023object}, extended the model to also function as a classifier \cite{li2021scouter,wang2023learning,wang2024explainable}, and incorporated pretrained encoders such as DINO \cite{caron2021emerging} \cite{seitzer2022bridging}.
While WWT borrows the concept of slots, existing methods all separate the encoder layers and the SA layers and connect them sequentially; WWT differs in that slots and masks are embedded in the backbone, formed in parallel with image features.

\paragraph{Multi-stream architecture}
Although near-single-path architectures dominate deep learning \cite{szegedy2015going,he2016deep}, multi-stream designs have been explored for specialized purposes:
ensembles for feature diversity and redundancy \cite{ciregan2012multi},
heterogeneous operators within a modality \cite{chen2017dual,chen2022mobile},
task-specialized pathways in multi-task learning \cite{misra2016cross,kawakami2019cross},
and multi-modal processing such as RGB and motion \cite{simonyan2014two}, image-text pairs \cite{radford2021learning}, or other sensor combinations \cite{xiao2020audiovisual,hao2024egocentric}.
WWT adopts a multi-stream structure, but derives two virtual modalities, semantics and position, from a single still image.
Unlike conventional multi-modal methods, these streams represent latent internal factors rather than external modalities.

\paragraph{What-where separation}
The idea of what-where separation has two independent origins: 
In neuroscience, the two-stream hypothesis \cite{mishkin1983object,milner1995visual} proposed distinct pathways for object recognition (``what'') and spatial processing (``where'' or ``how'').
This inspired two-stream CNNs modeling what-where separation through foveated or retinal transformations \cite{choi2023dual,ebrahimpour2019ventral}, but they relied on parallel CNN pathways rather than representational factorization.
Independently, early representation learning raised concerns that CNNs discard positional information through pooling \cite{hinton2011transforming}.
Subsequent models addressed this by encoding pose transformations \cite{sabour2017dynamic}, or preserving reference positions during pooling \cite{zhao2015stacked}.
Multi-object VAEs, precursors to object-centric learning, use attention maps to define responsibility regions for reconstruction \cite{burgess2019monet}; however, their latent variables remain global image-level vectors.
An orthogonal line of work has focused on improving positional embeddings and phase-based representations \cite{liu2025self,gopalakrishnan2025decoupling}.
WWT is the first ViT-style backbone that explicitly enforces what?where separation throughout the architecture.

\vspace{-5mm}\section{Method}\vspace{-1mm}
\vspace{-3mm}\subsection{What-Where Transformer: A Visual Backbone}\vspace{-3mm}
Given an input image, WWT outputs multiple pairs of representation vectors and localization masks corresponding to visual entities.
Let the token sequence length be denoted as $T = WH$.
We predefine the maximum number of visual entities as $S$,
and set this as the number of outputs. 
Each entity is assigned a $d$-dimensional vector called a slot, in line with object-centric-learning literature~\cite{locatello2020object}. 

\begin{figure}[t]
    \centering
    \hspace{-6mm}\includegraphics[width=1.0\linewidth]{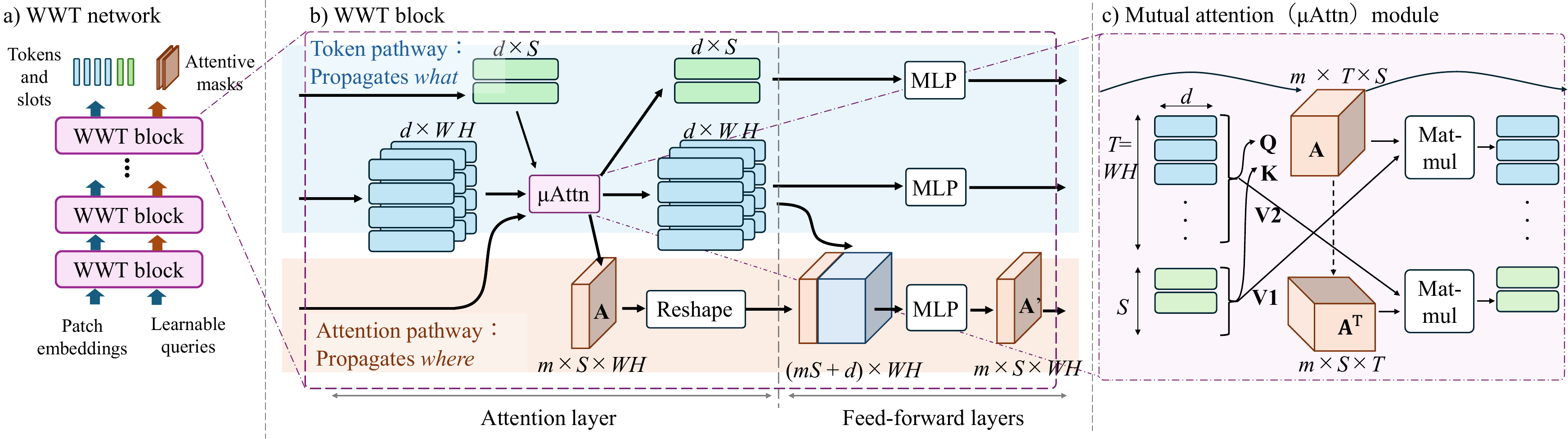}
     \vspace{-4mm}    
    \caption{The overview of WWT network, block, and mutual attention layer.}
    \vspace{-6mm}
    \label{fig:mattn}
\end{figure}

The network is designed to bind local patch tokens to slots to form meaningful groups during feedforward inference, while contextualizing tokens with slots as their summaries.
We therefore adopt slot-token mutual communication as the basic building block of the architecture.
A WWT block takes tokens, slots, and masks as inputs and outputs updated representations.

In the WWT block, we introduce a novel mutual attention layer ({$\mu$}Attn; Fig.~\ref{fig:mattn}), where tokens and slots interact via dot-product attention mechanism.
$\mu$Attn layer is defined using the token sequence $\bm{x}_i \in \mathbb{R}^{T \times d}$, the slot sequence $\bm{z}_i \in \mathbb{R}^{S \times d}$, and the mask $\bm{A}_i \in \mathbb{R}^{T \times S}$from the $i$-th layer, as follows:
\begin{eqnarray}
\bm{A}'_{i} &=& \bm{A}_{i} + \frac{Q(\bm{x}_i) K(\bm{z}_i)^\top}{\sqrt{d}}, \\
\bm{x}'_{i} &=& \bm{x}_{i} + \text{Softmax}_T(\bm{A'}_{i}) V_1(\bm{z}_{i}), \\
\bm{z}'_{i} &=& \bm{z}_{i} + \text{Softmax}_S (\bm{A'}^\top_{i})V_2(\bm{x}_{i}) ,
\end{eqnarray}

where $Q, K, V_1,$ and $V_2$ are projections that map tokens and slots into queries, keys, and values, and $\text{Softmax}_T$ and $\text{Softmax}_S$ denote softmax operations along the token and slot dimensions, respectively. This means token-to-slot and slot-to-token connections are symmetrically wired except for the normalization direction by softmax, to omit duplicated computing of attention.
Combining propagated masks with newly computed attention enables progressive refinement.
For simplicity, the above formulation is in the form of single-head attention; however, similar to conventional self- and cross-attention, it can be extended to a multi-head setting. When the number of heads is $m$, the dimensionality of tokens and slots remains unchanged, while only the mask becomes multi-channel, i.e., $\bm{A}_i \in \mathbb{R}^{m \times T \times S}$.

After the {$\mu$}Attn layer, MLP layers perform token-wise and slot-wise processing.
To improve the masks in the feedforward processing, we put point-wise MLP layers for the masks, too. 
Since the masks have a small number of channels and are not suitable for standalone MLP processing, they are concatenated with tokens along the channel axis before the MLP. The number of output channels is set to S, so that it can be reused as an attention map in the next layer.
These MLPs are denoted as
\begin{eqnarray}
\bm{x}_{i+1} &=& \bm{x}'_{i} + \mathrm{MLP}_T(\bm{x}'_{i}), \\
\bm{z}_{i+1} &=& \bm{z}'_{i} + \mathrm{MLP}_S(\bm{z}'_{i}), \\
\bm{A}_{i+1} &=& \mathrm{MLP}_A\big(\mathrm{Concat}(\bm{A}'_{i}, \bm{x}'_{i})\big),
\end{eqnarray}
where $\mathrm{MLP}_T$, $\mathrm{MLP}_S$, $\mathrm{MLP}_A$ are independent MLPs with unshared parameters. We use three MLPs per block but we modified the hidden-layer dimensionalities to have equivalent Flops to the ViT counterparts in total. 
Among them, $MLP_A$ constructs a novel design of MLP-over-attention, providing next blocks' learned attention biases in an input-conditioned manner. This is useful for eliminating spurious attentions on backgrounds and compensating vague attention boundaries by parametric feedforward computation.
Residual connections of $\mathrm{MLP}_A$ are removed due to training instability; raw residues seem to have too large values and possibly disturb attentions too much to provide sufficient convergence.

\begin{wrapfigure}{r}{0.36\textwidth}
    \centering
    \vspace{-5mm}\includegraphics[width=1.0\linewidth]{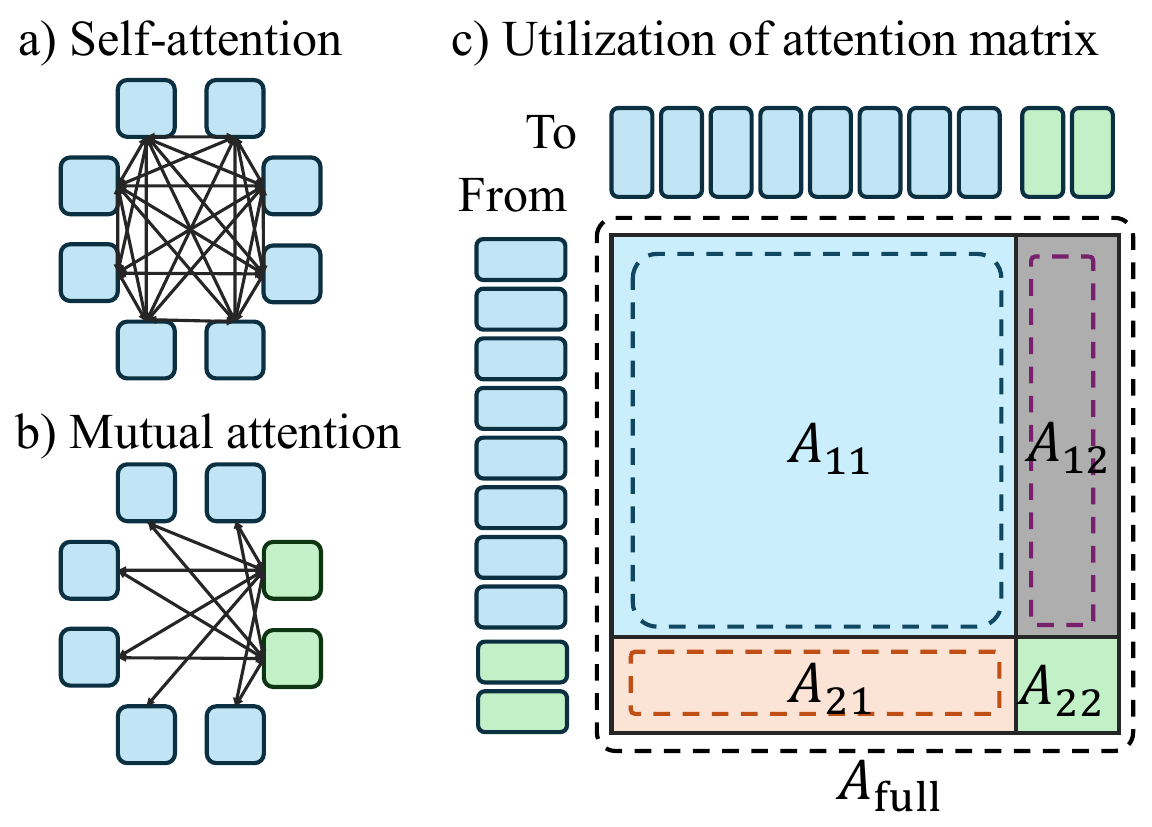}
  \vspace{-7mm} 
    \caption{Illustration of ViT's token-token connectivity and WWT's token-slot connectivity.}
    \vspace{-6mm}
    \label{fig:connect}
\end{wrapfigure}
The WWT network is configured as a stack of WWT blocks.
We omit spatial hierarchy or gradual downsampling for simplicity and to maintain the original-resolution attentions.
In the first WWT block of the network, the initial tokens are obtained by applying a linear transformation to the RGB values of each patch. The initial slots are learnable queries initialized randomly and updated during training \cite{carion2020end}. The initial mask is given as zero-filled values. Positional embeddings are added only to the initial tokens.
From the final $L$-th layer's outputs, $(\bm{z}_{L}, \bm{A}_{L}) = \mathbb{R}^{S \times (d + W \times H)}$ can be used as an image representation decomposed into $S$ factors.

\paragraph{Connectivity pattern} 
WWT removes token-to-token and slot-to-slot interactions, using only bidirectional slot-token communication.
Unlike self-attention, which mixes all tokens globally, WWT retains only the off-diagonal attention blocks $A_{12}$ and $A_{21}$ with a symmetry constraint $A_{12}=A_{21}^\top$ (Fig.~\ref{fig:connect}c).
This largely reduces computation and VRAM usage for high-resolution inputs, and is essential for efficiently implementing the attention pathway in WWT, which forwards $A_{21}$ across layers.
This connectivity bottleneck forces tokens to communicate only through slots, encouraging binding to informative slots for contextualization.

\paragraph{What-where decomposed representation} 
WWT outputs slot-mask pairs rather than dense feature tensors or token-sequence representations, yielding an explicitly decomposed representation. 
The slot-mask representation admits reconstruction into a dense feature tensor  as 
\begin{eqnarray}
F_{hwd}=\sum_{s=1}^{S}A_{shw}\bm{z}_{sd},
\end{eqnarray}
which can be interpreted as a learned factorization of dense feature maps into spatial assignment tensors and slot-wise feature bases. This formulation relates WWT to feature-factorization methods such as Deep Feature Factorization (DFF) \cite{collins2018deep,fel2023craft}, while differing in that WWT performs inference directly in the factorized latent space rather than decomposing features post hoc.
From the perspective of what-where separation, conventional feature maps encode {\it what} in feature vectors while relying on spatial coordinates as hard-coded {\it where} representations.
In contrast, WWT introduces explicit learnable {\it where} representations via slot-masks, potentially facilitating emergent grouping and non-local object-level interactions.

 \vspace{-3mm}\subsection{Task Heads Using Slots and Masks}\label{sec:heads} \vspace{-3mm}
 
\begin{figure}[t]
    \centering
    \includegraphics[width=1.0\linewidth]{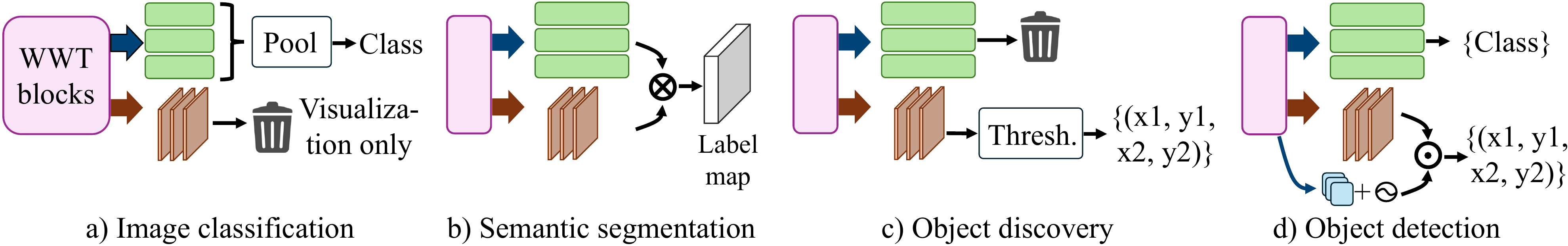}
     \vspace{-6mm}
    \caption{Task heads utilizing slot-mask representations from WWT (see Sec. \ref{sec:heads}).}
    \vspace{-3mm}
    \label{fig:task}
\end{figure}

By using the set of slot-mask pairs produced by WWT as an image representation, it can be applied to recognition and localization tasks with relatively simple task-specific heads as shown in Fig.~\ref{fig:task}.

\paragraph{Classification Head (a)} 
The output slots $\bm{z}_L \in \mathbb{R}^{ S \times d}$ are fed into a multilayer perceptron (MLP) and slot-wise $C$-class classification $\bm{c} \in \mathbb{R}^{S \times C}$ is obtained. After slot-wise classification, the results are averaged along the $S$ dimension to obtain the global image class.
Optionally, we can compute class activations (CA) for a predicted class $p$ from the slot-classes and output masks by a class-confidence-weights average of masks as
\begin{eqnarray}
    \text{CA} = \frac{1}{S}\sum_{s=1}^{S} \bm{c}(s, p) \text{ReLU}(A_L(s)) \in \mathbb{R}_{\geq0}^{W \times H},
\end{eqnarray}
where $\bm{c}(s, p)$ denotes the predicted probability (or logit) of class $p$ in slot $s$. The mask is allowed to have negative values, so we rectify them. This activation is used for visualization and interpretation.

\paragraph{Segmentation Head (b)} 
The output masks $\bm{A}_L$ are reassembled \cite{ranftl2021vision} into patch-level coarse region labels $\bm{M}_L \in \mathbb{R}^{W \times H \times S}$. If necessary, this is upsampled. This is then combined with per-slot classification results $\mathbb{R}^{C \times S}$ via matrix multiplication to obtain the final label map $\mathbb{R}^{C \times rW \times rH}$, where $r$ denotes the upsampling factor.
This head can be weakly supervised with image-level class labels by training the classification head and repurposing it, or supervised by dense pixel labels.

\paragraph{Object Discovery Head (c)}
In object discovery without additional training, object regions are obtained by binarizing each mask with a fixed threshold, extracting connected components, and removing duplicates.  
For single-object discovery, these are treated as candidates, and a post-processing step selects the most object-like region. Although heuristic, the region with the highest spatial concentration \cite{herfindahl1997concentration} over the mask is selected.

\paragraph{Object Detection Head (d)}
For box-supervised detection, WWT predicts boxes from masks in a differentiable manner.
To preserve spatial information, coordinate encodings \cite{liu2018intriguing} are concatenated to the patch tokens.
The mask is used to softly aggregate the tokens and coordinates in the object region, and the aggregated feature is fed into MLPs for box regression, as we found the slot embedding itself, focusing on {\it what}, to be insufficient for box regression.
Object classification, including the background class, is performed for each slot.
Similar to DETR, bipartite matching is used during training to assign slots to ground-truth objects.

\paragraph{Autoencoding Head} 
In addition to the task-specific heads, an autoencoding head is used to provide regularization during ImageNet training.  
Feature maps are obtained in a manner similar to the segmentation head, but inter-slot competition is introduced by applying softmax to $A_L$ along the slot axis $s$ before the multiplication.
A small CNN is trained to reconstruct the input RGB image from the feature map.
This is used as an auxiliary loss; through the autoencoding process, the masks $\bm{A}_L$ are encouraged to group regions that are easier to reconstruct together.
We also use this head for a distillation-based unsupervised object-centric-learning setting \cite{seitzer2022bridging}:
This setting is to ensure a fair comparison with DINOSAUR \cite{seitzer2022bridging}.
Instead of RGB values, the head predicts a feature map produced by a frozen self-supervised DINO backbone \cite{caron2021emerging} via an MLP decoder.

\begin{table}[t]
\vspace{-3mm}
  \caption{Classification accuracy and explainability on ImageNet-1k.}
  \label{tab:imagenet}
  \centering
  \begin{tabular}{lccccc}
    \toprule
    \multicolumn{3}{c}{Model Size} & Accuracy & \multicolumn{2}{c}{Explainability} \\
    \cmidrule(r){1-3} \cmidrule(l){5-6}
    Model & FLOPs & Params & Top-1 Acc. & Drop$\downarrow$ & Inc.$\uparrow$ \\
    \midrule
    DeiT-Tiny        & 1.3G & 5M  & \textbf{72.2}\% & 52.6\% & 0.5\% \\
    \hspace{3mm}+ Rollout & --   & --  & --            & 38.6\% & 4.4\% \\
    WWT-Tiny  (ours)    & 1.3G & 9M  & 71.3\%          & \textbf{30.1}\% & \textbf{6.8}\% \\
    \midrule
    DeiT-Small       & 4.6G & 22M & \textbf{79.9}\% & 57.5\% & 0.2\% \\
    WWT-Small  (ours)   & 4.8G & 34M & 76.1\%          & \textbf{44.0}\% & \textbf{4.8}\% \\
    \bottomrule
  \end{tabular}
  \vspace{-3mm}
\end{table}
\begin{table}
\vspace{-3mm}
\caption{Weakly Supervised Semantic Segmentation Performance (mIoU) on ImageNet-S 50}
\vspace{-0mm}
\label{tab:imagenets}
\centering
\begin{tabular}{lll}
\toprule
Method & Activation & mIoU (\%) \\
\midrule
DeiT-Tiny & Raw CLS attention & 8.8 \\
DeiT-Tiny & Rollout \cite{abnar2020quantifying} & 35.9 \\
WWT-Tiny (ours) & Raw mask output & \textbf{49.1} \\
\midrule
ResNet18 \cite{he2016deep} & SEAM \cite{wang2020self} & 45.2 \\
(reprod. by \cite{gao2022large}) & SC-CAM \cite{chang2020weakly} & 38.5 \\
 & AdvCAM \cite{lee2021anti} & 46.9 \\
\bottomrule
\end{tabular}
\vspace{-6mm}
\end{table}

\vspace{-4mm}\subsection{Relationship with Existing Architecture}\vspace{-2mm}
\paragraph{ViT}
WWT extends ViT by replacing token-wise self-attention with slot-based mutual attention, while explicitly factorizing representations into semantic (what) and positional (where) streams.
This design enables object-level representation with reduced complexity by decoupling patch length $T$ and slot length $S$.
Unlike ViT, which relies on a single CLS token, WWT maintains multiple slots that act similarly to registers \cite{darcet2024vision} but are explicitly tied to object-level representations, contributing to the emergence of object masks.

\paragraph{DETR}
WWT shares the use of learned queries and cross-attention-like interactions with DETR, but embeds these mechanisms within the encoder.
This design unifies feature extraction and localization without requiring a separate decoder, reinforcing a backbone-centric formulation for localization.
In contrast to DETR, where objectness does not naturally emerge without strong supervision, even in self-supervised settings \cite{dai2021up,chen2023siamese}, but WWT exhibits object-centric representations as an inherent property of the architecture.

\paragraph{Slot Attention}
WWT incorporates slot-based object representations directly into the backbone, eliminating the need for a separate slot attention module.
In contrast to conventional SA, which relies on iterative slot binding, WWT amortizes the recurrent binding process into a non-iterative backbone forward pass, enabling joint feature learning and slot formation.

\vspace{-3mm}\section{Experiments}\vspace{-3mm}\
We investigated the performance of WWT on various tasks involving natural images by conducting zero-shot transfer without additional training, and unsupervised and supervised finetuning after pretraining on ImageNet-1k.

\paragraph{Implementation and training}
We follow DeiT \cite{touvron2021training} a supervised ViT training recipe in ImageNet, with only minor modifications.
We prepared two model variants: WWT-Tiny and WWT-Small comparable with the equivalent-sized ViTs.
Pre-training was conducted on the ImageNet-1k training set with both a classification head and a self-encoding head attached. The classification head was trained using cross-entropy loss with label smoothing ($p = 0.1$) \cite{szegedy2016rethinking}, while the autoencoding head was trained using mean squared error loss on normalized RGB input values. Gradients were backpropagated through the entire network via both the slots and the masks.
The ImageNet-pretrained WWT-Small was finetuned into WWT-OCL for unsupervised discovery, WWT-DET and WWT-SEG for supervised detection and segmentation.
Further details are provided in the Appendix.

\paragraph{Classification and explainability}
First, we verified whether the ImageNet supervised pretraining was successful based on the classification results in Table~\ref{tab:imagenet} and the explainability metrics.
In terms of classification accuracy, our method achieves performance comparable to DeiT~\cite{touvron2021training}, 
and although it is slightly inferior for larger models, it is rather promising that comparable accuracy is maintained even when all fully connected self-attention across tokens is replaced with sparser mutual attention.

\begin{table}[t]
\centering
\vspace{-3mm}
\caption{CorLoc (\%) for zero-shot single object discovery}
\vspace{-0mm}
\small
\label{tab:lost}
\begin{tabular}{lcccc}
\toprule
Model & Training & Post-processing & VOC12 & COCO20k \\
\midrule
DeiT (reprod. by \cite{darcet2024vision}) & Supervised & LOST & 11.7\% & 10.7\% \\
DeiT + Reg.~\cite{darcet2024vision} & Supervised & LOST & 27.1\% & 25.1\% \\
DeiT + LaSt~\cite{shi2026vision} (our reprod.) & Supervised & Threshold & 29.7\% & 23.1\% \\ 
WWT (ours) & Supervised & Threshold & \textbf{41.4\%} & \textbf{30.2\%} \\
\midrule
\textcolor{gray}{Obj-DINOv2 \cite{rawlekar2026finding}} & \textcolor{gray}{Self-supervised} & \textcolor{gray}{TokenCut} & \textcolor{gray}{30.7\%} & \textcolor{gray}{19.7\%} \\
\textcolor{gray}{Obj-DINOv3 \cite{rawlekar2026finding}} & \textcolor{gray}{Self-supervised} & \textcolor{gray}{TokenCut} & \textcolor{gray}{36.0\%} & \textcolor{gray}{23.4\%} \\
\midrule
\textcolor{gray}{DINOv2 (reprod. by \cite{darcet2024vision})} & \textcolor{gray}{Self-supervised} & \textcolor{gray}{LOST} & \textcolor{gray}{35.3\%} & \textcolor{gray}{26.9\%} \\
\textcolor{gray}{DINOv2 + Reg.~\cite{darcet2024vision}} & \textcolor{gray}{Self-supervised} & \textcolor{gray}{LOST} & \textcolor{gray}{55.4\%} & \textcolor{gray}{42.0\%} \\
\textcolor{gray}{DINOv1} & \textcolor{gray}{Self-supervised} & \textcolor{gray}{LOST} & \textcolor{gray}{\textbf{61.9\%}} & \textcolor{gray}{\textbf{50.7\%}} \\
\bottomrule
\end{tabular}
\vspace{-0mm}
\end{table}
\begin{table}
\vspace{-6mm}
\caption{Recall metrics for zero-shot multi-object discovery}
\vspace{-0mm}
\small
\label{tab:most}
\centering
\begin{tabular}{llllll}
\toprule
\multicolumn{2}{c}{Method} & \multicolumn{2}{c}{VOC12} & \multicolumn{2}{c}{COCO20k} \\
\cmidrule(r){1-2} \cmidrule(r){3-4} \cmidrule(r){5-6}
Model Name & Post-processing & \#Predictions & Recall & \#Predictions & Recall \\
\midrule
WWT (ours) & Threshold & 16.3 & \textbf{42.4}\% & 16.2 & \textbf{19.8}\% \\
DINOv1 & MOST(1) \cite{rambhatla2023most} & 3.0 & 33.9\% & 4.9 & 12.4\% \\
DINOv1 & MOST(2) & 7.7 & 34.1\% & 11.5 & 13.7\% \\
DINOv1 & MOST(3) & 12.5 & 34.1\% & 14.8 & 13.6\% \\
\bottomrule
\end{tabular}
\vspace{-6mm}
\end{table}

Furthermore, regarding the Drop and Increase (Inc.) metrics \cite{chattopadhay2018grad}, which measure the validity of input attribution for classification results, WWT significantly outperforms DeiT when using class-token attention maps or attention masks as they are. This advantage is not reversed even when the post-hoc interpretation method Rollout \cite{abnar2020quantifying} is applied only to ViT.

\begin{figure*}[t]
    \centering
    \includegraphics[width=.98\linewidth]{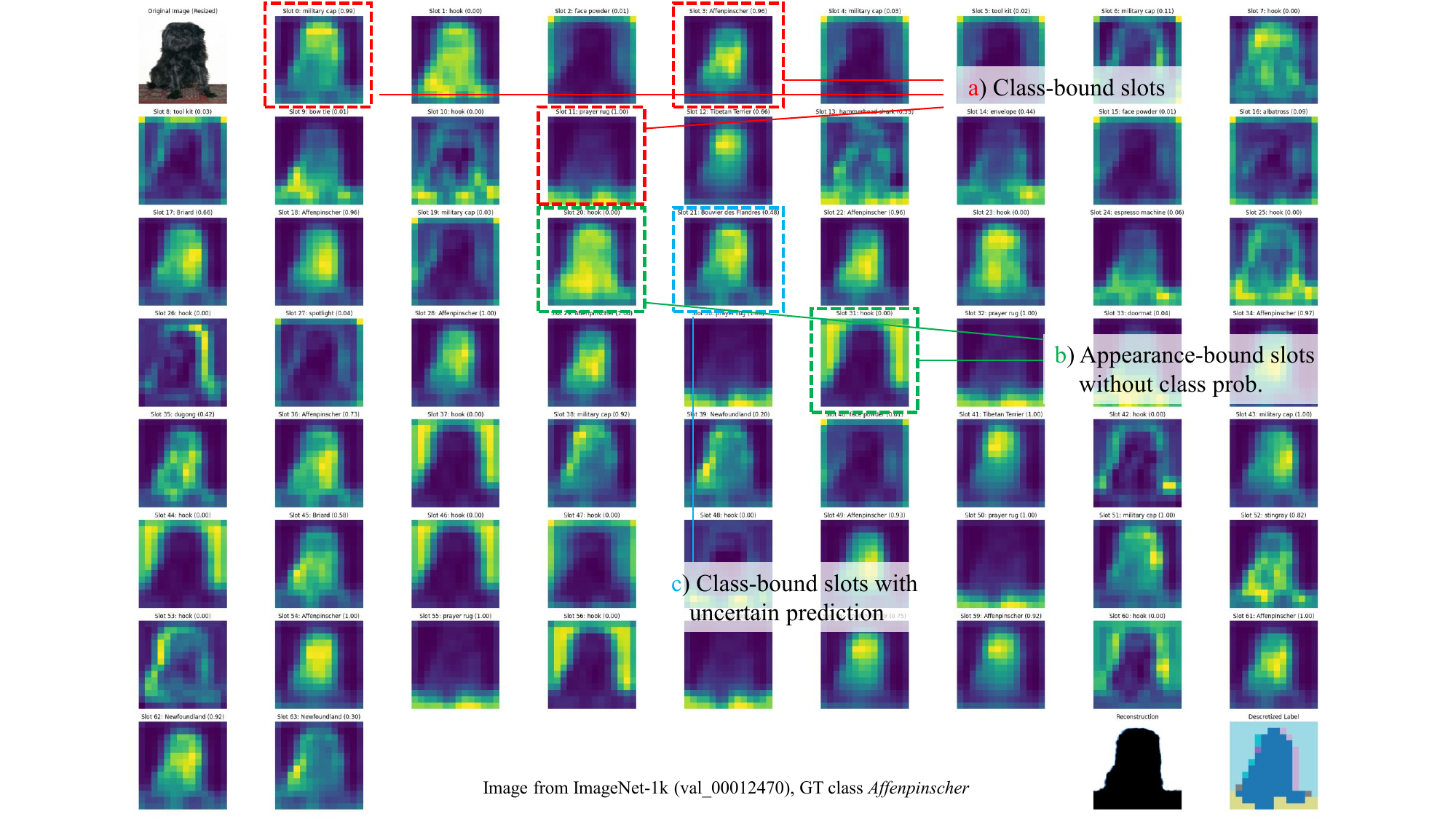}
    \vspace{-3mm}
    \caption{Visualization of per-slot output masks.
Red (a): Class-bound slots attend to semantically meaningful regions and predict classes corresponding to the attended regions (with occasional prediction errors).
Blue (b): Appearance-bound slots attend to regions but assign negligible class probabilities, e.g., {\it hook} with 0.0\% top-1 class probability. Green (c): Class-bound slots with uncertainty attend to relevant regions but produce low-confidence predictions over confusing classes.}
    \vspace{-0mm}
    \label{fig:slot1}
\end{figure*}

\paragraph{ImageNet weakly supervised segmentation}
To evaluate whether the attention maps capture object regions more precisely, we conducted a weakly supervised segmentation protocol using a classifier on the region-labeled subset ImageNet-S \cite{gao2022large} 50. As shown in Table~\ref{tab:imagenets}, WWT again demonstrated superior performance compared to DeiT and the baseline methods provided by the dataset authors. Notably, WWT outperformed convolutional ResNet \cite{he2016deep}, which is considered to have more local sensitivity than attentive models, combined with CAM-specific training methods \cite{wang2020self,chang2020weakly,lee2021anti}.

\paragraph{Zero-shot object discovery}
Zero-shot object discovery \cite{simeoni2021localizing,rambhatla2023most}, which applies models to detection benchmarks without further training, was then conducted. Following the standard protocol, class labels were ignored, and evaluation was performed using CorLoc \cite{gokberk2014multi}, which does not penalize detections of non-annotated objects, as well as recall.
While ViTs can usually perform object discovery via token clustering, e.g., LOST \cite{simeoni2021localizing} and MOST \cite{rambhatla2023most}, WWT can be applied simply by thresholding its masks. Even in single-object discovery, WWT outperformed DeiT, including variants with registers, and achieved accuracy comparable to some of the strong self-supervised methods using  DINOv2 \cite{oquab2024dinov2} and v3 (Table \ref{tab:lost}) \footnote{DINOv1’s strong performance in object discovery, unmatched by v2 and v3, has been reported in the literature \cite{darcet2024vision,rawlekar2026finding}.}.

In multi-object discovery (Table \ref{tab:most}), WWT exhibited a larger number of discoveries and higher recall than DINOv1-based MOST. Although recall tends to increase as more predictions are produced by definition, increasing the clustering granularity in DINO+MOST to generate as many predictions as WWT leads to over-segmentation, causing recall to plateau. These results suggest that WWT’s capability for multi-object discovery is not a trivial property of ViT-based approaches.

\begin{table}
\vspace{-7mm}
\caption{Unsupervised object discovery results in the DINOSAUR \cite{seitzer2022bridging} setting.}
\vspace{-0mm}
\label{tab:ocl}
\centering
\small
\begin{tabular}{lccccc}
\toprule
 & & \multicolumn{2}{c}{VOC12} & \multicolumn{2
}{c}{COCO} \\
\cmidrule(r){3-4} \cmidrule(r){5-6}
Method & Distil.& mBO$_i$ & mBO$_c$ & mBO$_i$ & mBO$_c$ \\
\midrule
WWT-OCL, zero-shot & & 29.9 & 31.8  & 16.2 &  19.5  \\
WWT-OCL, RGB & & 22.7 & 25.0  & 14.9 & 16.3  \\
WWT-OCL & \checkmark & \underline{37.5 $\pm$ 0.2} & \underline{39.8 $\pm$ 0.2} & \underline{24.0 $\pm$ 0.3} & \underline{27.4 $\pm$ 0.2}  \\
\midrule
\hspace{-2mm} {\it With MLP decoders} & & && &\\
Slot attention & & 24.6 & 24.9 & 17.2 & 19.2  \\
DINOSAUR-MLP & \checkmark & \textbf{39.5 $\pm$ 0.1} & \textbf{40.9$\pm$ 0.1} & \textbf{27.7$\pm$0.2} & \textbf{30.9$\pm$0.2}  \\
\hspace{-2mm} {\it \textcolor{gray}{With autoregressive decoders}} & & & & &\\ 
\textcolor{gray}{SLATE} && \textcolor{gray}{31.0 $\pm$ 0.4} &  \textcolor{gray}{32.4 $\pm$ 0.4}& -- & -- \\
\textcolor{gray}{DINOSAUR-Trans.} & \checkmark &  \textcolor{gray}{44.0 $\pm$ 0.8} & \textcolor{gray}{51.2 $\pm$ 0.8} & \textcolor{gray}{32.3$\pm$0.4} &   \textcolor{gray}{38.8$\pm$0.4} \\
\bottomrule
\end{tabular}
\vspace{-6mm}
\end{table}

\begin{figure*}[t]
    \centering
    \includegraphics[width=0.965\linewidth]{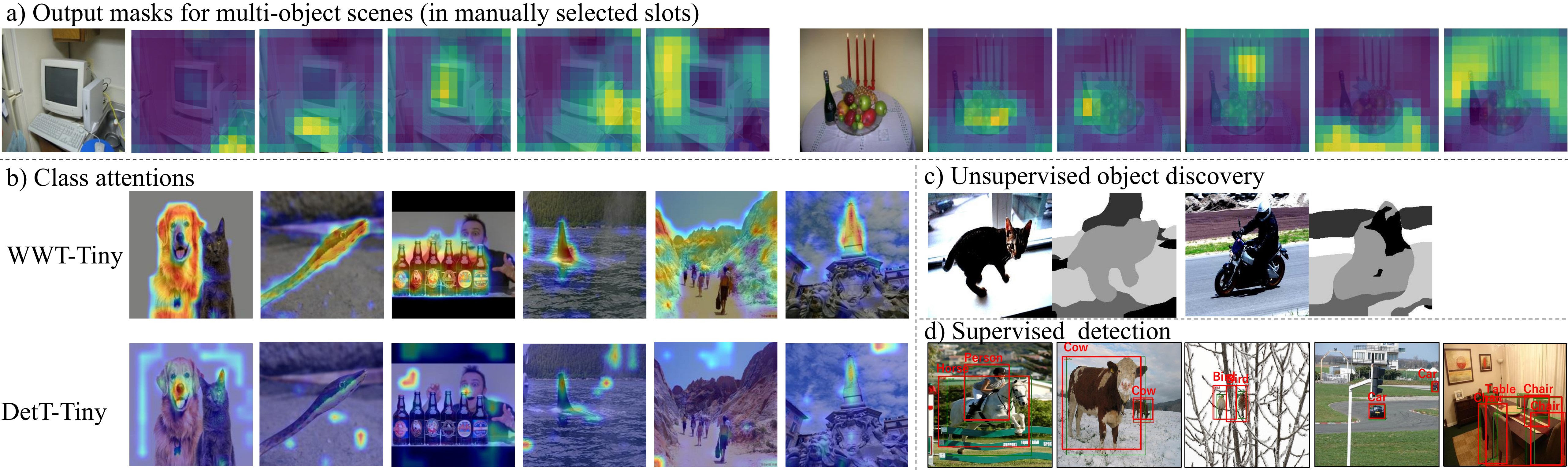}
    \vspace{-2mm}
    \caption{Visualization of the WWT's attentive masks and task results.}
    \vspace{-5mm}
    \label{fig:vis}
\end{figure*}

\paragraph{Unsupervised object discovery}
Many object-centric methods address unsupervised object discovery through slot-based autoencoding \cite{locatello2020object,seitzer2022bridging,kakogeorgiou2024spot}.
For comparison, we finetuned WWT under the same setting.
WWT uses 64 backbone slots, far more than the six or seven slots typically used in SA-based models for VOC and COCO. During finetuning, reconstruction was applied only to the first six or seven slot-mask pairs, while the remaining slots were discarded.
\begin{wrapfigure}{r}{0.40\textwidth}
    \centering
    \vspace{-4mm}\includegraphics[width=1.0\linewidth]{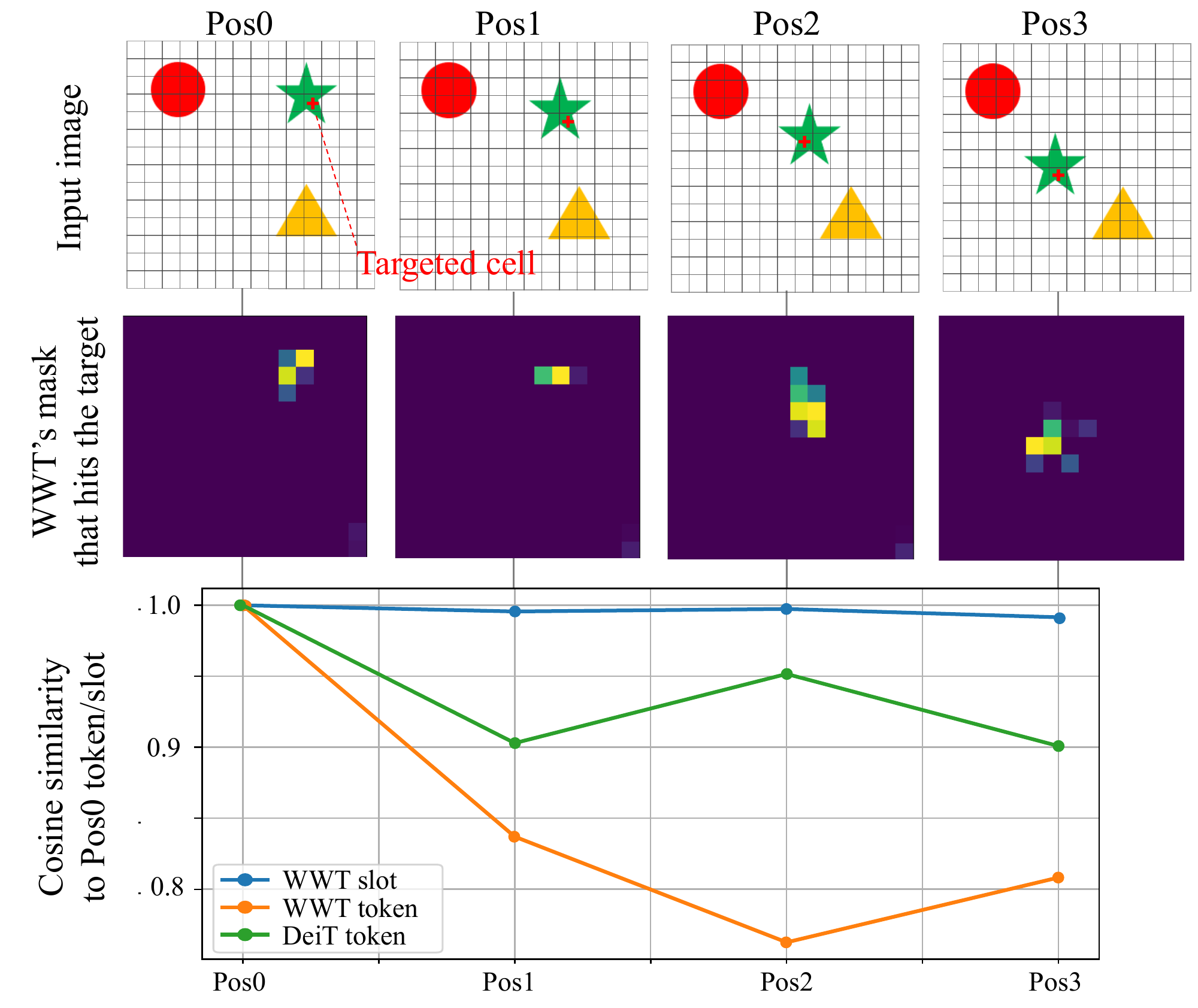}
    \vspace{-6mm}   
    \caption{ Translation invariance of patch tokens and slots.}
    \vspace{-8mm}
    \label{fig:wwsep}
\end{wrapfigure}
The results in Tab. \ref{tab:ocl} suggest that WWT performed mask-based discovery comparably with the SA-based baseline method, i.e., DINOSAUR, in the distillation-based setting.
While some SA-based methods benefit from stronger autoregressive Transformer decoders, WWT enables object centricity within the backbone itself. Its ability to perform this task without additional SA modules suggests that WWT internalized the representational bottleneck required for object binding.
We also found that distillation from self-supervised dense representations is inevitable; RGB-reconstruction-based finetuning using the smaller datasets degraded the discovery performances from the zero-shot application of ImageNet-pretrained weights.

\paragraph{Analyzing what-where separation}
As an analytical evidence of what-where separation in WWT, we plotted the translation invariance of patch tokens and slots in Fig. \ref{fig:wwsep} using artificially placed patterns.
We gathered patch tokens' and slots' embeddings that cover a target pattern, moving from Pos0 to Pos3,
and plotted the similarity of each embedding to Pos0's.
WWT's slot showed strong translation invariance, while tokens were variable along the movements.
This suggests the existence of what-where separation, where slots carry translation-invariant {\it what} and the factors of translation are visible in {\it where} masks.
More examples are in the Appendix.

\paragraph{Supervised proving in detection and segmentation}
WWT is readily applicable to supervised tasks with only minor modifications, owing to its backbone-inbuilt localization mechanism, unlike prior zero-shot and unsupervised localization methods relying on non-differentiable modules \cite{simeoni2021localizing,wang2022self}.
As a demonstration of this, we conducted supervised finetuning for bounding-box detection and semantic segmentation. Table \ref{tab:supervised} summarizes the results, 
showing the feasibility of supervised finetuning with minimalistic task heads.
Moreover, the proposed task-head settings in WWT-DET and WWT-SEG outperformed their heads-on-patch-tokens counterparts, suggesting the emergence of role separation in the decomposed representation; at least some part of information in tokens was transported to slots and masks, and as a result, solving these tasks solely by tokens became suboptimal. The results of downstream tasks are visualized in Fig. \ref{fig:vis}.

\paragraph{Hyperparameter and ablations}
Table \ref{tab:ablation} summarizes results under various hyperparameter settings and ablations. Varying the number of slots shows that 64 achieves a good trade-off between accuracy and computational cost: increasing the number yields no further gains, while reducing it degrades performance. Notably, although ImageNet images contain far fewer than 64 objects, this indicates that the required number of slots is driven more by representational capacity than object count. The ablation study also confirms the effectiveness of using an MLP over attention and the autoencoding head. We note that these experiments were conducted under a preliminary augmentation setting, which leads to slight differences in numbers from the main results.

\begin{table}
\vspace{-3mm}
\caption{Supervised proving results in detection and semantic segmentation in VOC}
\vspace{-0mm}
\label{tab:supervised}
\centering
\small
\begin{tabular}{llcc}
\toprule
Detection in VOC07 & mAP (\%) \\
  &  \\
\midrule
WWT-DET & 63.1 \\
WWT + det. head on tokens &  37.8 \\
\midrule
DETR & 72.2 \\
\bottomrule
\end{tabular} \hspace{2mm}
\begin{tabular}{llcc}
\toprule
Segmentation in VOC12 & mIoU (\%) \\
 &  \\
\midrule
WWT-SEG & 61.3 \\
WWT + seg. head on tokens & 59.9 \\
\midrule
ResNet-50 FCN  & 66.5 \\
\bottomrule
\end{tabular}
\vspace{-8mm} 
\end{table}

\begin{table} 
 \vspace{-8mm} 
\caption{Hyperparameter and ablative analyses.} 
\vspace{-0mm} 
\label{tab:ablation} 
\small \centering 
\begin{tabular}{lcccccc} 
\toprule
& & Full & & W/o MLP over attention  & W/o AE head \\
\#Slots & $S=48$ & $S=64$ & $S=96$ & $S=64$ & $S=64$ \\ 
\midrule
Top-1 acc. & 66.8 & 69
7 & 69.6 & 67.8 & 69.6\\ 
CorLoc & -- & 38.7 & --& 23.4& 19.8\\ \bottomrule 
\end{tabular} 
\vspace{-5mm} 
\end{table}

\vspace{-6mm} \section{Conclusion}\vspace{-5mm} 
We proposed a novel attentive vision architecture WWT, which decomposed the semantic and positional information of objects into slots and masks.
By introducing the inductive bias of what?where separation, WWT learned object-focused attention masks even under simple classification settings.
This design enabled the model to handle positional information as a distinct and generalizable component of image representation, without relying on specific decoders for downstream tasks. We believe that task-generic localization within the encoder can benefit object grounding and spatial reasoning by mitigating annotation-dependency, which would be important in future VLM/VLA systems. We hope this work serves as a starting point for future studies on object-centric visual backbones.
\paragraph{Limitations}
Our focus in this work is on classification-based supervised pretraining.
Investigating whether WWT maintains the emergent object-decomposing property in self-supervised or text-contrastive training settings is an open and interesting future research direction.

\begin{ack}
This work was supported by DENSO IT LAB Recognition, Control, and Learning Algorithm Collaborative Research Chair. The experiments were carried out using the TSUBAME4.0 supercomputer at Institute of Science Tokyo.
\end{ack}

{
\small
\bibliographystyle{plain}
\bibliography{myref} 

@string{NIPS  = "Advances in Neural Information Processing Systems"}

@string{ICCV  = "Proceedings of the IEEE/CVF International Conference on Computer Vision"}

@string{CVPR  = "Proceedings of the IEEE/CVF Conference on Computer Vision and Pattern Recognition"}

@string{CVPRW  = "Proceedings of the IEEE/CVF Conference on Computer Vision and Pattern Recognition Workshops"}

@string{ECCV  = "European Conference on Computer Vision"}

@string{ICML  = "International Conference on Machine Learning"}

@string{ICLR  = "International Conference on Learning Representations"}

@string{ICLRW  = "International Conference on Learning Representations Workshops"}

@string{TPAMI = "IEEE Transactions on Pattern Analysis and Machine Intelligence"}

@string{ACL   = "Annual Meeting of the Association for Computational Linguistics"}

@string{WACV  = "Winter Conference on Applications of Computer Vision"}

@string{ARXIV = "arXiv preprint"}

@inproceedings{deng2009imagenet,
  title={{ImageNet}: A large-scale hierarchical image database},
  author={Deng, Jia and Dong, Wei and Socher, Richard and Li, Li-Jia and Li, Kai and Fei-Fei, Li},
  booktitle=CVPR,
  pages={248--255},
  year={2009},
  organization={Ieee}
}

@article{gao2022large,
  title={Large-scale unsupervised semantic segmentation},
  author={Gao, Shanghua and Li, Zhong-Yu and Yang, Ming-Hsuan and Cheng, Ming-Ming and Han, Junwei and Torr, Philip},
  journal=TPAMI,
  volume={45},
  number={6},
  pages={7457--7476},
  year={2022},
  publisher={IEEE}
}

@article{vaswani2017attention,
  title={Attention is all you need},
  author={Vaswani, Ashish and Shazeer, Noam and Parmar, Niki and Uszkoreit, Jakob and Jones, Llion and Gomez, Aidan N and Kaiser, {\L}ukasz and Polosukhin, Illia},
  journal=NIPS,
  volume={30},
  year={2017}
}

@article{liu2018intriguing,
  title={An intriguing failing of convolutional neural networks and the {CoordConv} solution},
  author={Liu, Rosanne and Lehman, Joel and Molino, Piero and Petroski Such, Felipe and Frank, Eric and Sergeev, Alex and Yosinski, Jason},
  journal=NIPS,
  volume={31},
  year={2018}
}

@inproceedings{dosovitskiy2020image,
  title={An Image is Worth 16x16 Words: Transformers for Image Recognition at Scale},
  author={Dosovitskiy, Alexey and Beyer, Lucas and Kolesnikov, Alexander and Weissenborn, Dirk and Zhai, Xiaohua and Unterthiner, Thomas and Dehghani, Mostafa and Minderer, Matthias and Heigold, Georg and Gelly, Sylvain and others},
  booktitle=ICLR
}

@inproceedings{touvron2021training,
  title={Training data-efficient image transformers \& distillation through attention},
  author={Touvron, Hugo and Cord, Matthieu and Douze, Matthijs and Massa, Francisco and Sablayrolles, Alexandre and J{\'e}gou, Herv{\'e}},
  booktitle=ICML,
  pages={10347--10357},
  year={2021}
}

@inproceedings{caron2021emerging,
  title={Emerging properties in self-supervised vision transformers},
  author={Caron, Mathilde and Touvron, Hugo and Misra, Ishan and J{\'e}gou, Herv{\'e} and Mairal, Julien and Bojanowski, Piotr and Joulin, Armand},
  booktitle=ICCV,
  pages={9650--9660},
  year={2021}
}

@inproceedings{wang2021pyramid,
  title={Pyramid vision transformer: A versatile backbone for dense prediction without convolutions},
  author={Wang, Wenhai and Xie, Enze and Li, Xiang and Fan, Deng-Ping and Song, Kaitao and Liang, Ding and Lu, Tong and Luo, Ping and Shao, Ling},
  booktitle=ICCV,
  pages={568--578},
  year={2021}
}

@inproceedings{wu2021cvt,
  title={{CvT}: Introducing convolutions to vision transformers},
  author={Wu, Haiping and Xiao, Bin and Codella, Noel and Liu, Mengchen and Dai, Xiyang and Yuan, Lu and Zhang, Lei},
  booktitle=ICCV,
  pages={22--31},
  year={2021}
}

@inproceedings{liu2021swin,
  title={Swin transformer: Hierarchical vision transformer using shifted windows},
  author={Liu, Ze and Lin, Yutong and Cao, Yue and Hu, Han and Wei, Yixuan and Zhang, Zheng and Lin, Stephen and Guo, Baining},
  booktitle=ICCV,
  pages={10012--10022},
  year={2021}
}

@inproceedings{ranftl2021vision,
  title={Vision transformers for dense prediction},
  author={Ranftl, Ren{\'e} and Bochkovskiy, Alexey and Koltun, Vladlen},
  booktitle=ICCV,
  pages={12179--12188},
  year={2021}
}

@article{liu2021pay,
  title={Pay attention to mlps},
  author={Liu, Hanxiao and Dai, Zihang and So, David and Le, Quoc V},
  journal=NIPS,
  volume={34},
  pages={9204--9215},
  year={2021}
}

@article{touvron2022resmlp,
  title={Resmlp: Feedforward networks for image classification with data-efficient training},
  author={Touvron, Hugo and Bojanowski, Piotr and Caron, Mathilde and Cord, Matthieu and El-Nouby, Alaaeldin and Grave, Edouard and Izacard, Gautier and Joulin, Armand and Synnaeve, Gabriel and Verbeek, Jakob and others},
  journal=TPAMI,
  volume={45},
  number={4},
  pages={5314--5321},
  year={2022},
  publisher={IEEE}
}

@inproceedings{chen2022mobile,
  title={Mobile-former: Bridging mobilenet and transformer},
  author={Chen, Yinpeng and Dai, Xiyang and Chen, Dongdong and Liu, Mengchen and Dong, Xiaoyi and Yuan, Lu and Liu, Zicheng},
  booktitle=CVPR,
  pages={5270--5279},
  year={2022}
}

@inproceedings{liu2022convnet,
  title={A convnet for the 2020s},
  author={Liu, Zhuang and Mao, Hanzi and Wu, Chao-Yuan and Feichtenhofer, Christoph and Darrell, Trevor and Xie, Saining},
  booktitle=CVPR,
  pages={11976--11986},
  year={2022}
}

@article{yu2023metaformer,
  title={Metaformer baselines for vision},
  author={Yu, Weihao and Si, Chenyang and Zhou, Pan and Luo, Mi and Zhou, Yichen and Feng, Jiashi and Yan, Shuicheng and Wang, Xinchao},
  journal=TPAMI,
  volume={46},
  number={2},
  pages={896--912},
  year={2023},
  publisher={IEEE}
}

@article{ma2023image,
  title={Image as set of points},
  author={Ma, Xu and Zhou, Yuqian and Wang, Huan and Qin, Can and Sun, Bin and Liu, Chang and Fu, Yun},
  journal=ICLR,
  year={2023}
}

@article{yao2023dual,
  title={Dual vision transformer},
  author={Yao, Ting and Li, Yehao and Pan, Yingwei and Wang, Yu and Zhang, Xiao-Ping and Mei, Tao},
  journal=TPAMI,
  volume={45},
  number={9},
  pages={10870--10882},
  year={2023}
}

@inproceedings{dehghani2023scaling,
  title={Scaling vision transformers to 22 billion parameters},
  author={Dehghani, Mostafa and Djolonga, Josip and Mustafa, Basil and Padlewski, Piotr and Heek, Jonathan and Gilmer, Justin and Steiner, Andreas Peter and Caron, Mathilde and Geirhos, Robert and Alabdulmohsin, Ibrahim and others},
  booktitle=ICML,
  pages={7480--7512},
  year={2023}
}

@article{deng2023perceptual,
  title={Perceptual group tokenizer: Building perception with iterative grouping},
  author={Deng, Zhiwei and Chen, Ting and Li, Yang},
  journal=ICLR,
  year={2024}
}

@article{oquab2024dinov2,
  title={{DINOv2}: Learning Robust Visual Features without Supervision},
  author={Oquab, Maxime and Darcet, Timoth{\'e}e and Moutakanni, Th{\'e}o and Vo, Huy and Szafraniec, Marc and Khalidov, Vasil and Fernandez, Pierre and Haziza, Daniel and Massa, Francisco and El-Nouby, Alaaeldin and others},
  journal={Transactions on Machine Learning Research Journal},
  year={2024}
}

@inproceedings{darcet2024vision,
  title={Vision Transformers Need Registers},
  author={Darcet, Timoth{\'e}e and Oquab, Maxime and Mairal, Julien and Bojanowski, Piotr},
  booktitle=ICLR,
  year={2024}
}

@article{shi2026vision,
  title={Vision Transformers Need More Than Registers},
  author={Shi, Cheng and Yu, Yizhou and Yang, Sibei},
  journal=CVPR,
  year={2026}
}

@inproceedings{li2025token,
  title={Token activation map to visually explain multimodal llms},
  author={Li, Yi and Wang, Hualiang and Ding, Xinpeng and Wang, Haonan and Li, Xiaomeng},
  booktitle=CVPR,
  pages={48--58},
  year={2025}
}

@inproceedings{minderer2022simple,
  title={Simple open-vocabulary object detection},
  author={Minderer, Matthias and Gritsenko, Alexey and Stone, Austin and Neumann, Maxim and Weissenborn, Dirk and Dosovitskiy, Alexey and Mahendran, Aravindh and Arnab, Anurag and Dehghani, Mostafa and Shen, Zhuoran and others},
  booktitle=ECCV,
  pages={728--755},
  year={2022},
  organization={Springer}
}

@inproceedings{li2022exploring,
  title={Exploring plain vision transformer backbones for object detection},
  author={Li, Yanghao and Mao, Hanzi and Girshick, Ross and He, Kaiming},
  booktitle=ECCV,
  pages={280--296},
  year={2022}
}

@inproceedings{carion2020end,
  title={End-to-end object detection with transformers},
  author={Carion, Nicolas and Massa, Francisco and Synnaeve, Gabriel and Usunier, Nicolas and Kirillov, Alexander and Zagoruyko, Sergey},
  booktitle=ECCV,
  pages={213--229},
  year={2020}
}

@article{zhu2020deformable,
  title={Deformable {DETR}: Deformable transformers for end-to-end object detection},
  author={Zhu, Xizhou and Su, Weijie and Lu, Lewei and Li, Bin and Wang, Xiaogang and Dai, Jifeng},
  journal=ICLR,
  year={2021}
}

@article{cheng2021per,
  title={Per-pixel classification is not all you need for semantic segmentation},
  author={Cheng, Bowen and Schwing, Alex and Kirillov, Alexander},
  journal=NIPS,
  volume={34},
  pages={17864--17875},
  year={2021}
}

@inproceedings{dai2021up,
  title={{UP-DETR}: Unsupervised pre-training for object detection with transformers},
  author={Dai, Zhigang and Cai, Bolun and Lin, Yugeng and Chen, Junying},
  booktitle=CVPR,
  pages={1601--1610},
  year={2021}
}

@inproceedings{chen2023siamese,
  title={Siamese {DETR}},
  author={Chen, Zeren and Huang, Gengshi and Li, Wei and Teng, Jianing and Wang, Kun and Shao, Jing and Loy, Chen Change and Sheng, Lu},
  booktitle=CVPR,
  pages={15722--15731},
  year={2023}
}

@inproceedings{cheng2022masked,
  title={Masked-attention mask transformer for universal image segmentation},
  author={Cheng, Bowen and Misra, Ishan and Schwing, Alexander G and Kirillov, Alexander and Girdhar, Rohit},
  booktitle=CVPR,
  pages={1290--1299},
  year={2022}
}

@inproceedings{zhang2022dino,
  title={{DINO}: {DETR} with improved denoising anchor boxes for end-to-end object detection},
  author={Zhang, Hao and Li, Feng and Liu, Shilong and Zhang, Lei and Su, Hang and Zhu, Jun and Ni, Lionel M and Shum, Heung-Yeung},
  booktitle=ICLR,
  year={2023}
}

@inproceedings{kirillov2023segment,
  title={Segment anything},
  author={Kirillov, Alexander and Mintun, Eric and Ravi, Nikhila and Mao, Hanzi and Rolland, Chloe and Gustafson, Laura and Xiao, Tete and Whitehead, Spencer and Berg, Alexander C and Lo, Wan-Yen and others},
  booktitle=ICCV,
  pages={4015--4026},
  year={2023}
}

@article{burgess2019monet,
  title={{NONet}: Unsupervised scene decomposition and representation},
  author={Burgess, Christopher P and Matthey, Loic and Watters, Nicholas and Kabra, Rishabh and Higgins, Irina and Botvinick, Matt and Lerchner, Alexander},
  journal={arXiv preprint arXiv:1901.11390},
  year={2019}
}

@article{locatello2020object,
  title={Object-centric learning with slot attention},
  author={Locatello, Francesco and Weissenborn, Dirk and Unterthiner, Thomas and Mahendran, Aravindh and Heigold, Georg and Uszkoreit, Jakob and Dosovitskiy, Alexey and Kipf, Thomas},
  journal=NIPS,
  volume={33},
  pages={11525--11538},
  year={2020}
}

@inproceedings{singh2021illiterate,
  title={Illiterate {DALL-E} learns to compose},
  author={Singh, Gautam and Deng, Fei and Ahn, Sungjin},
  booktitle=ICLR,
  year={2022}
}

@inproceedings{li2021scouter,
  title={Scouter: Slot attention-based classifier for explainable image recognition},
  author={Li, Liangzhi and Wang, Bowen and Verma, Manisha and Nakashima, Yuta and Kawasaki, Ryo and Nagahara, Hajime},
  booktitle=ICCV,
  pages={1046--1055},
  year={2021}
}

@inproceedings{wu2023inverted,
  title={Inverted-attention transformers can learn object representations: Insights from slot attention},
  author={Wu, Yi-Fu and Greff, Klaus and Elsayed, Gamaleldin Fathy and Mozer, Michael Curtis and Kipf, Thomas and van Steenkiste, Sjoerd},
  booktitle={Causal Representation Learning Workshop at NeurIPS},
  year={2023}
}

@inproceedings{wang2023learning,
  title={Learning bottleneck concepts in image classification},
  author={Wang, Bowen and Li, Liangzhi and Nakashima, Yuta and Nagahara, Hajime},
  booktitle=CVPR,
  pages={10962--10971},
  year={2023}
}

@article{wang2024explainable,
  title={Explainable Image Recognition via Enhanced Slot-attention Based Classifier},
  author={Wang, Bowen and Li, Liangzhi and Zhang, Jiahao and Nakashima, Yuta and Nagahara, Hajime},
  journal={arXiv preprint arXiv:2407.05616},
  year={2024}
}

@inproceedings{kakogeorgiou2024spot,
  title={{SPOT}: Self-training with patch-order permutation for object-centric learning with autoregressive transformers},
  author={Kakogeorgiou, Ioannis and Gidaris, Spyros and Karantzalos, Konstantinos and Komodakis, Nikos},
  booktitle=CVPR,
  pages={22776--22786},
  year={2024}
}

@article{liu2025self,
  title={Self-Supervised Learning of Intertwined Content and Positional Features for Object Detection},
  author={Liu, Kang Jun and Suganuma, Masanori and Okatani, Takayuki},
  journal=ICML,
  volume={267},
  pages={39552--39567},
  year={2025},
  publisher={ML Research Press}
}

@article{gopalakrishnan2025decoupling,
  title={Decoupling the ``What'' and ``Where'' With Polar Coordinate Positional Embeddings},
  author={Gopalakrishnan, Anand and Csord{\'a}s, Robert and Schmidhuber, J{\"u}rgen and Mozer, Michael C},
  journal={arXiv preprint arXiv:2509.10534},
  year={2025}
}

@article{chung2014empirical,
  title={Empirical evaluation of gated recurrent neural networks on sequence modeling},
  author={Chung, Junyoung and Gulcehre, Caglar and Cho, KyungHyun and Bengio, Yoshua},
  journal=ARXIV,
  volume={arXiv:1412.3555},
  year={2014}
}

@article{seitzer2022bridging,
  title={Bridging the gap to real-world object-centric learning},
  author={Seitzer, Maximilian and others},
  journal=ICLR,
  year={2023}
}

@article{jiang2023object,
  title={Object-centric slot diffusion},
  author={Jiang, Jindong and Deng, Fei and Singh, Gautam and Ahn, Sungjin},
  journal=NIPS,
  volume={arXiv:2303.10834},
  year={2023}
}

@inproceedings{chattopadhay2018grad,
  title={Grad-{CAM}++: Generalized gradient-based visual explanations for deep convolutional networks},
  author={Chattopadhay, Aditya and Sarkar, Anirban and Howlader, Prantik and Balasubramanian, Vineeth N},
  booktitle=WACV,
  pages={839--847},
  year={2018},
  organization={IEEE}
}

@inproceedings{collins2018deep,
  title={Deep feature factorization for concept discovery},
  author={Collins, Edo and Achanta, Radhakrishna and Susstrunk, Sabine},
  booktitle=ECCV,
  pages={336--352},
  year={2018}
}

@inproceedings{abnar2020quantifying,
  title={Quantifying attention flow in transformers},
  author={Abnar, Samira and Zuidema, Willem},
  booktitle=ACL,
  pages={4190--4197},
  year={2020}
}

@inproceedings{wang2020self,
  title={Self-supervised equivariant attention mechanism for weakly supervised semantic segmentation},
  author={Wang, Yude and Zhang, Jie and Kan, Meina and Shan, Shiguang and Chen, Xilin},
  booktitle=CVPR,
  pages={12275--12284},
  year={2020}
}

@inproceedings{chang2020weakly,
  title={Weakly-supervised semantic segmentation via sub-category exploration},
  author={Chang, Yu-Ting and Wang, Qiaosong and Hung, Wei-Chih and Piramuthu, Robinson and Tsai, Yi-Hsuan and Yang, Ming-Hsuan},
  booktitle=CVPR,
  pages={8991--9000},
  year={2020}
}

@inproceedings{lee2021anti,
  title={Anti-adversarially manipulated attributions for weakly and semi-supervised semantic segmentation},
  author={Lee, Jungbeom and Kim, Eunji and Yoon, Sungroh},
  booktitle=CVPR,
  year={2021}
}

@inproceedings{xu2022multi,
  title={Multi-class token transformer for weakly supervised semantic segmentation},
  author={Xu, Lian and Ouyang, Wanli and Bennamoun, Mohammed and Boussaid, Farid and Xu, Dan},
  booktitle=CVPR,
  pages={4310--4319},
  year={2022}
}

@inproceedings{gupta2022vitol,
  title={{ViTOL}: Vision transformer for weakly supervised object localization},
  author={Gupta, Saurav and Lakhotia, Sourav and Rawat, Abhay and Tallamraju, Rahul},
  booktitle=CVPRW,
  pages={4101--4110},
  year={2022}
}

@inproceedings{fel2023craft,
  title={{CRAFT}: Concept recursive activation factorization for explainability},
  author={Fel, Thomas and Picard, Agustin and Bethune, Louis and Boissin, Thibaut and Vigouroux, David and Colin, Julien and Cad{\`e}ne, R{\'e}mi and Serre, Thomas},
  booktitle=CVPR,
  pages={2711--2721},
  year={2023}
}

@inproceedings{gokberk2014multi,
  title={Multi-fold {MIL} training for weakly supervised object localization},
  author={Gokberk Cinbis, Ramazan and Verbeek, Jakob and Schmid, Cordelia},
  booktitle=CVPR,
  pages={2409--2416},
  year={2014}
}

@inproceedings{simeoni2021localizing,
  title={Localizing Objects with Self-Supervised Transformers and no Labels},
  author={Sim{\'e}oni, Oriane and Puy, Gilles and Vo, Huy V and Roburin, Simon and Gidaris, Spyros and Bursuc, Andrei and P{\'e}rez, Patrick and Marlet, Renaud and Ponce, Jean},
  booktitle=BMVC,
  year={2021}
}

@inproceedings{wang2022self,
  title={Self-supervised transformers for unsupervised object discovery using normalized cut},
  author={Wang, Yangtao and Shen, Xi and Hu, Shell Xu and Yuan, Yuan and Crowley, James L and Vaufreydaz, Dominique},
  booktitle=CVPR,
  pages={14543--14553},
  year={2022}
}

@inproceedings{rambhatla2023most,
  title={{MOST}: Multiple object localization with self-supervised transformers for object discovery},
  author={Rambhatla, Sai Saketh and Misra, Ishan and Chellappa, Rama and Shrivastava, Abhinav},
  booktitle=ICCV,
  pages={15823--15834},
  year={2023}
}

@inproceedings{psomas2023keep,
  title={Keep it {SimPool}: Who said supervised transformers suffer from attention deficit?},
  author={Psomas, Bill and Kakogeorgiou, Ioannis and Karantzalos, Konstantinos and Avrithis, Yannis},
  booktitle=ICCV,
  pages={5350--5360},
  year={2023}
}

@inproceedings{rawlekar2026finding,
  title={Finding Distributed Object-Centric Properties in Self-Supervised Transformers},
  author={Rawlekar, Samyak and Swain, Amitabh and Cai, Yujun and Wang, Yiwei and Yang, Ming-Hsuan and Ahuja, Narendra},
  booktitle=CVPR,
  year={2026}
}

@inproceedings{ciregan2012multi,
  title={Multi-column deep neural networks for image classification},
  author={Ciregan, Dan and Meier, Ueli and Schmidhuber, J{\"u}rgen},
  booktitle=CVPR,
  pages={3642--3649},
  year={2012},
  organization={IEEE}
}

@article{simonyan2014two,
  title={Two-stream convolutional networks for action recognition in videos},
  author={Simonyan, Karen and Zisserman, Andrew},
  journal=NIPS,
  volume={27},
  year={2014}
}

@inproceedings{misra2016cross,
  title={Cross-stitch networks for multi-task learning},
  author={Misra, Ishan and Shrivastava, Abhinav and Gupta, Abhinav and Hebert, Martial},
  booktitle=CVPR,
  pages={3994--4003},
  year={2016}
}

@inproceedings{kawakami2019cross,
  title={Cross-connected networks for multi-task learning of detection and segmentation},
  author={Kawakami, Rei and Yoshihashi, Ryota and Fukuda, Seiichiro and You, Shaodi and Iida, Makoto and Naemura, Takeshi},
  booktitle=ICIP,
  pages={3636--3640},
  year={2019},
  organization={IEEE}
}

@inproceedings{ebrahimpour2019ventral,
  title={Ventral-dorsal neural networks: object detection via selective attention},
  author={Ebrahimpour, Mohammad K and Li, Jiayun and Yu, Yen-Yun and Reesee, Jackson and Moghtaderi, Azadeh and Yang, Ming-Hsuan and Noelle, David C},
  booktitle=WACV,
  pages={986--994},
  year={2019},
  organization={IEEE}
}

@article{xiao2020audiovisual,
  title={Audiovisual slowfast networks for video recognition},
  author={Xiao, Fanyi and Lee, Yong Jae and Grauman, Kristen and Malik, Jitendra and Feichtenhofer, Christoph},
  journal={arXiv preprint arXiv:2001.08740},
  year={2020}
}

@inproceedings{radford2021learning,
  title={Learning transferable visual models from natural language supervision},
  author={Radford, Alec and Kim, Jong Wook and Hallacy, Chris and Ramesh, Aditya and Goh, Gabriel and Agarwal, Sandhini and Sastry, Girish and Askell, Amanda and Mishkin, Pamela and Clark, Jack and others},
  booktitle=ICML,
  pages={8748--8763},
  year={2021},
  organization={PmLR}
}

@inproceedings{choi2023dual,
  title={A Dual-Stream Neural Network Explains the Functional Segregation of Dorsal and Ventral Visual Pathways in Human Brains},
  author={Choi, Minkyu and Han, Kuan and Wang, Xiaokai and Zhang, Yizhen and Liu, Zhongming},
  booktitle=NIPS,
  year={2023}
}

@article{hao2024egocentric,
  title={Egocentric human activities recognition with multimodal interaction sensing},
  author={Hao, Yuzhe and Kanezaki, Asako and Sato, Ikuro and Kawakami, Rei and Shinoda, Koichi},
  journal={IEEE Sensors Journal},
  volume={24},
  number={5},
  pages={7085--7096},
  year={2024},
  publisher={IEEE}
}

@article{chen2017dual,
  title={Dual path networks},
  author={Chen, Yunpeng and Li, Jianan and Xiao, Huaxin and Jin, Xiaojie and Yan, Shuicheng and Feng, Jiashi},
  journal=NIPS,
  volume={30},
  year={2017}
}

@inproceedings{he2016deep,
  title={Deep residual learning for image recognition},
  author={He, Kaiming and Zhang, Xiangyu and Ren, Shaoqing and Sun, Jian},
  booktitle=CVPR,
  pages={770--778},
  year={2016}
}

@inproceedings{szegedy2016rethinking,
  title={Rethinking the inception architecture for computer vision},
  author={Szegedy, Christian and Vanhoucke, Vincent and Ioffe, Sergey and Shlens, Jon and Wojna, Zbigniew},
  booktitle=CVPR,
  pages={2818--2826},
  year={2016}
}

@inproceedings{szegedy2015going,
  title={Going deeper with convolutions},
  author={Szegedy, Christian and Liu, Wei and Jia, Yangqing and Sermanet, Pierre and Reed, Scott and Anguelov, Dragomir and Erhan, Dumitru and Vanhoucke, Vincent and Rabinovich, Andrew},
  booktitle=CVPR,
  pages={1--9},
  year={2015}
}

@article{mishkin1983object,
  title={Object vision and spatial vision: two cortical pathways},
  author={Mishkin, Mortimer and Ungerleider, Leslie G and Macko, Kathleen A},
  journal={Trends in neurosciences},
  volume={6},
  pages={414--417},
  year={1983},
  publisher={Elsevier}
}

@book{milner1995visual,
  title={The Visual Brain in Action},
  author={Milner, A. David and Goodale, Melvyn A.},
  year={1995},
  publisher={Oxford University Press},
  address={Oxford}
}

@book{herfindahl1997concentration,
  title={Concentration in the steel industry},
  author={Herfindahl, Orris C},
  year={1997},
  publisher={Columbia university}
}

@article{gregor2010emergence,
  title={Emergence of complex-like cells in a temporal product network with local receptive fields},
  author={Gregor, Karo and LeCun, Yann},
  journal={arXiv preprint arXiv:1006.0448},
  year={2010}
}

@inproceedings{hinton2011transforming,
  title={Transforming auto-encoders},
  author={Hinton, Geoffrey E and Krizhevsky, Alex and Wang, Sida D},
  booktitle={International conference on artificial neural networks},
  pages={44--51},
  year={2011},
  organization={Springer}
}

@inproceedings{sabour2017dynamic,
  title={Dynamic Routing Between Capsules},
  author={Sabour, Sara and Frosst, Nicholas and Hinton, Geoffrey E},
  booktitle=NIPS,
  year={2017}
}

@inproceedings{zhao2015stacked,
  title={Stacked what-where auto-encoders},
  author={Zhao, Junbo and Mathieu, Michael and Goroshin, Ross and Lecun, Yann},
  booktitle=ICLRW,
  year={2016}
}

@article{goyal2022inductive,
  title={Inductive biases for deep learning of higher-level cognition},
  author={Goyal, Anirudh and Bengio, Yoshua},
  journal={Proceedings of the Royal Society A},
  volume={478},
  number={2266},
  pages={20210068},
  year={2022},
  publisher={The Royal Society}
}

@article{kimura2024permutation,
  title={On permutation-invariant neural networks},
  author={Kimura, Masanari and Shimizu, Ryotaro and Hirakawa, Yuki and Goto, Ryosuke and Saito, Yuki},
  journal={arXiv preprint arXiv:2403.17410},
  year={2024}
}
}

\end{document}